\documentclass[runningheads]{llncs}
\usepackage{graphicx}
\usepackage{cite}
\usepackage{varioref}     
\usepackage{float} 

\usepackage{array}
\newcolumntype{P}[1]{>{\centering\arraybackslash}p{#1}}

\begin{document}

\title{Process Outcome Prediction: \\CNN vs. LSTM (with Attention)}
%
%
\author{Hans Weytjens\and
Jochen De Weerdt}
\authorrunning{H. Weytjens, J. De Weerdt}
%
\institute{Research Centre for Information Systems Engineering (LIRIS),\\ KU Leuven, Leuven, Belgium\\
\email{\{hans.weytjens,jochen.deweerdt\}@kuleuven.be}}
\maketitle              
\begin{abstract}
The early outcome prediction of ongoing or completed processes confers competitive advantage to organizations. The performance of classic machine learning and, more recently, deep learning techniques such as Long Short-Term Memory (LSTM) on this type of classification problem has been thorougly investigated. Recently, much research focused on applying Convolutional Neural Networks (CNN) to time series problems including classification, however not yet to outcome prediction. The purpose of this paper is to close this gap and compare CNNs to LSTMs. Attention is another technique that, in combination with LSTMs, has found application in time series classification and was included in our research. Our findings show that all these neural networks achieve satisfactory to high predictive power provided sufficiently large datasets. CNNs perfom on par with LSTMs; the Attention mechanism adds no value to the latter. Since CNNs run one order of magnitude faster than both types of LSTM, their use is preferable. All models are robust with respect to their hyperparameters and achieve their maximal predictive power early on in the cases, usually after only a few events, making them highly suitable for runtime predictions. We argue that CNNs' speed, early predictive power and robustness should pave the way for their application in process outcome prediction.

\end{abstract}
%
%
%
\section{Introduction}
\label{sec:intro}
 \vspace*{-2mm}
Every organization will gain considerable advantage from the early outcome prediction of its ongoing processes: they can attempt to influence undesired outcomes for the better, make well-informed decisions further down the value chain, or realize efficiency gains. Process mining is the exercise of extracting information from event logs stored in computer systems with the aim of discovering or improving them. Within this realm, predictive process monitoring focuses on making predictions and has already applied a plethora of machine learning approaches, achieving varying degrees of success. 
Teinemaa et al. \cite{teinemaa} benchmark several classical approaches such as random forests, gradient boosted trees (both based on decision trees), logistic regression and support vector machines using various datasets. These classical machine techniques sometimes rely heavily on manual feature engineering to represent the data, which is far from a trivial task. Deep learning techniques have enjoyed remarkable successes automatically representing data as a hierarchy of useful features. This led to a growing body of predictive process monitoring applications. Both Evermann et al. \cite{evermann} and Tax et al. \cite{tax} use Long Short-Term Memory neural networks (LSTMs) to predict next events and time stamps in business processes. Camargo et al. \cite{camargo} further refined this approach. Hinkka et al. \cite{Outcome_RNN} were amongst the first to apply them in process outcome prediction. Kratsch et al. \cite{DL_for_outcome_pred} included LSTM in their comprehensive comparison of deep learning and classical approaches for outcome prediction. LSTMs sometimes suffer from their limited memory capacity, an issue that Bahdanau et al. \cite{bahd} address with the Attention mechanism. Wang et al. \cite{bidir_att} used LSTMs with Attention in their outcome prediction benchmarking study that also included bidirectional LSTMs and classic approaches. Convolutional Neural Networks (CNNs) work with fixed-size, spatially-organized data and are often associated with computer vision. Nevertheless, one-dimensional CNNs are also utilized for time series classification or sequence modeling \cite{bai}. Fawaz et al.'s \cite{tsc} large-scale empirical study of deep learning methods for time series classification includes CNNs but not LSTMs.  Pasqualdibisceglie et al. \cite{2d}, however, opted for an original two-dimensional CNN approach to predict next events in processes. 
 
 \vspace*{-2mm}
\begin{table}
\centering

\caption{Related research in process mining and positioning of this paper.}\label{tab:rr}
\scalebox{0.95}{
\begin{tabular}{|p{2.5cm}|P{1.5cm}|P{1.5cm}|P{1.5cm}|P{1.5cm}|P{1.5cm}|P{1.5cm}|}
\hline
Paper &  Classic & LSTM & LSTM & CNN & Outcome & Early\\
& Machine Learning & & Attention & & Prediction & Prediction\\
\hline
Teinemaa \cite{teinemaa}&x&-&-&-&x&x\\ 
Evermann \cite{evermann}&-&x&-&-&-&-\\ 
Tax \cite{tax}&-&x&-&-&-&x\\ 
Camargo  \cite{camargo}&-&x&-&-&-&-\\ 
Hinkka \cite{Outcome_RNN}&-&x&-&-&x&x\\ 
Kratsch  \cite{DL_for_outcome_pred}&x&x&-&-&x&x\\ 
Wang  \cite{bidir_att}&x&x&x&-&x&x\\ 
Pasqual.  \cite{2d}&-&x&-&2D&-&-\\ 
\hline
This paper&-&x&x&x&x&x\\ 
\hline

\end{tabular}
}
 \vspace*{-2mm}
\end{table}

To the best of our knowledge, there is no published research investigating CNNs in process outcome prediction. Table~\ref{tab:rr} relates our paper to the research described above. It concentrates on comparing CNNs to LSTMs with and without Attention. We not only investigate classifying completed processes, but also look at predicting the outcomes of ongoing processes.  We find that neural networks can make accurate and early predictions, provided the datasets are large enough. CNNs are much faster than LSTMs, but deliver very similar results. Attention does not improve the plain-vanilla LSTMs. All three models prove to be robust, to be relatively insensitive to hyperparameter changes. Intriguingly, the time-related features and even the ordering of the events seem to play a minor role at best for the quality of the models.

\section{Solving the learning problem}
\vspace*{-2mm}
Our objective, to benchmark different neural networks against each other, guided the methodological choices described in this section. The choices we made will not necessarily be optimal for any given learner on any given dataset. All datasets are event logs describing processes, often called cases or traces. These processes consist of events. A number of attributes, also called features or variables, describe the events. Every case is associated with a binary outcome, also called class or target, e.g. `approved' vs. `non-approved' in the case of a loan application process.  In this paper, we use the words `cases', `events',  `features' and `targets'. The word `prefix' refers to ongoing, incomplete cases.

The learning problem is essentially to train a learner using a training dataset containing events, described by their features and organized in cases that are labeled with targets, with the goal of predicting the targets of unseen cases (complete or ongoing).

 \vspace*{-1mm}
\subsection{Models}

Recurrent Neural Networks (RNN) are neural networks specifically designed to handle sequences of variable length. Since process mining datasets are organized in cases containing a variable number of events that can be chronologically sorted using their respective time stamps, RNNs are intuitively the first choice when applying neural networks to process prediction problems as testified by their extensive use in the form of LSTMs (the most prominent RNNs, specifically designed to treat longer-term dependencies) in \cite{evermann,tax,camargo}. 

\begin{figure}[H]
\vspace*{-5mm}
\centering
  \begin{tabular}{@{}c@{}} 
    \includegraphics[scale = .90]{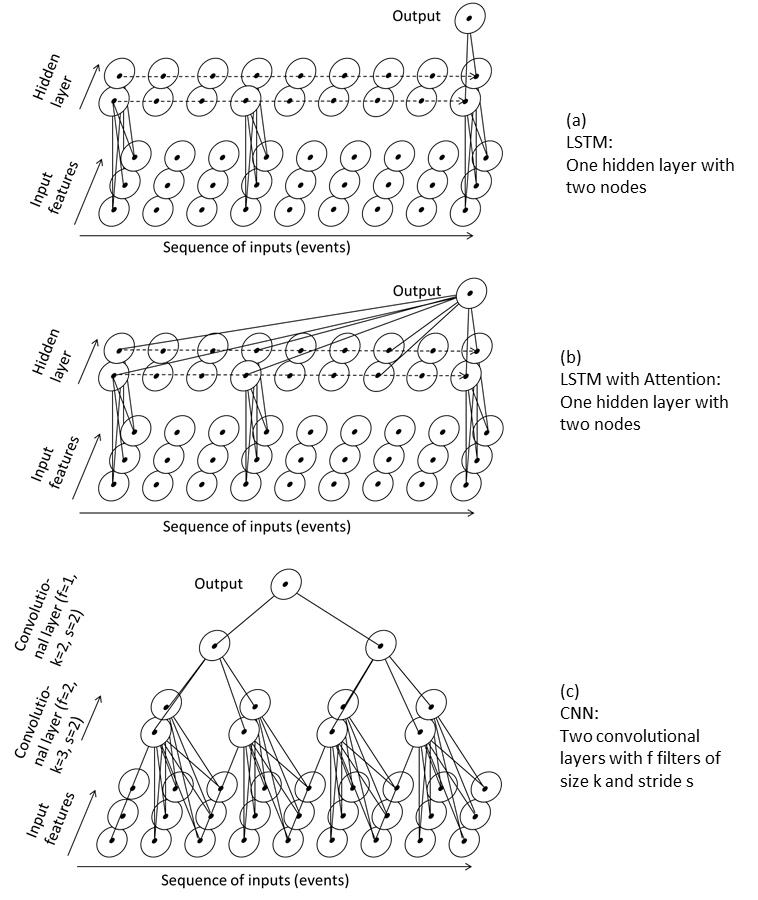} \\
   \vspace*{-5mm}
  \end{tabular}
 
  \caption{Conceptual Visualization Models (dense layers at top omitted).}\label{fig:models}
\vspace*{-4mm}
\end{figure}

An LSTM processes every sequence of events it is presented one time step at a time. At any given time step, it will pass a vector (aka `state of the memory cell') containing information about the current and previous time steps to the next time step, until reaching the last one (as depicted by the dotted horizontal arrows in Fig~\ref{fig:models}(a)) whose output is propagated to the next layer. The vector's fixed size, however, will inevitably limit its informational content. Especially for longer sequences, the information from the earlier inputs (events) risks dilution or loss. The Attention Mechanism shown in Fig~\ref{fig:models}(b) was proposed to overcome this problem by retaining the outputs of all nodes in the hidden layer, scoring them, and then calculating a weighted average of the outputs using these scores to compute the final outcome of the model (lines from the nodes in the hidden layer to the output) .

In contrast to LSTMs, Convolutional Neural Networks work with fixed-sized, spatially-organized data. A series of alternating convolution layers applying weight-sharing filters and dimension-reducing pooling layers enables the models to automatically recognize patterns and extract features from the input data. These features are then passed to a series of dense layers for classification (or regression).  Two-dimensional CNNs are very commonly used in computer vision applications. Interpreting time as a spatial dimension, one-dimensional CNNs can be applied to sequence processing as well. Fig~\ref{fig:models}(c) shows such a 1-D CNN with the filters striding along the temporal axis.

 \vspace*{-1mm}
\subsection{Preprocessing}\label{ss:pp}
The data we used and the targets (outcomes that are based on certain events in the cases) we defined are described in Section~\ref{sec: eval}. We labeled the cases by adding a target column to the original datasets and then clipped every case just before the event indicating its target value. To improve comparability, possibly at the detriment of the final result, we decided against incorporating any human domain knowledge. We made an exception for the following synthetic features, the same for all datasets, which were calculated as shown in Table~\ref{tab:sf}.

Since our datasets contain both range (e.g. `nr\_open') and categorical features (e.g. `month'), we proceeded to map the labels of the categorical features to integers. All our models include an embedding layer, which maps these integer values into vectors that should ideally be similar for values with comparable properties. The length of these vectors was set to be one-fifth of the respective feature's vocabulary size (nearly always far below 10, rarely above), thereby realizing a substantial dimensionality reduction compared to one-hot encoding. The embeddings themselves are learned by the model. Finally, we also standardized (mean of zero and standard deviation of one) the range features. 

\begin{table}[htp]
\centering
\vspace*{-3mm}
\caption{The synthetic features.}\label{tab:sf}
\begin{tabular}{|l|l|l|}
\hline
Name &  Type & Explanation\\
\hline
nr\_open & range & nr. of open cases at time of every event's time stamp (=load) \\
elapsed & range & time elapsed since start of case, marked by its first event \\
evTime & range & time since last event (0 for first event in the case) \\
sinceMidnight &  range & time elapsed since midnight of previous day \\
month & categ. & month of year \\
day & categ. & day of month \\
hour & categ. & hour of day \\
evNR & range & order nr of event in case\\ 
\hline

\end{tabular}
\vspace*{-5mm}
\end{table}

 \vspace*{-0mm}
\subsection{Feeding the models}\label{ss:fm}
\begin{figure}[H]
\vspace*{-2mm}
\center
\includegraphics[width=.6\textwidth]{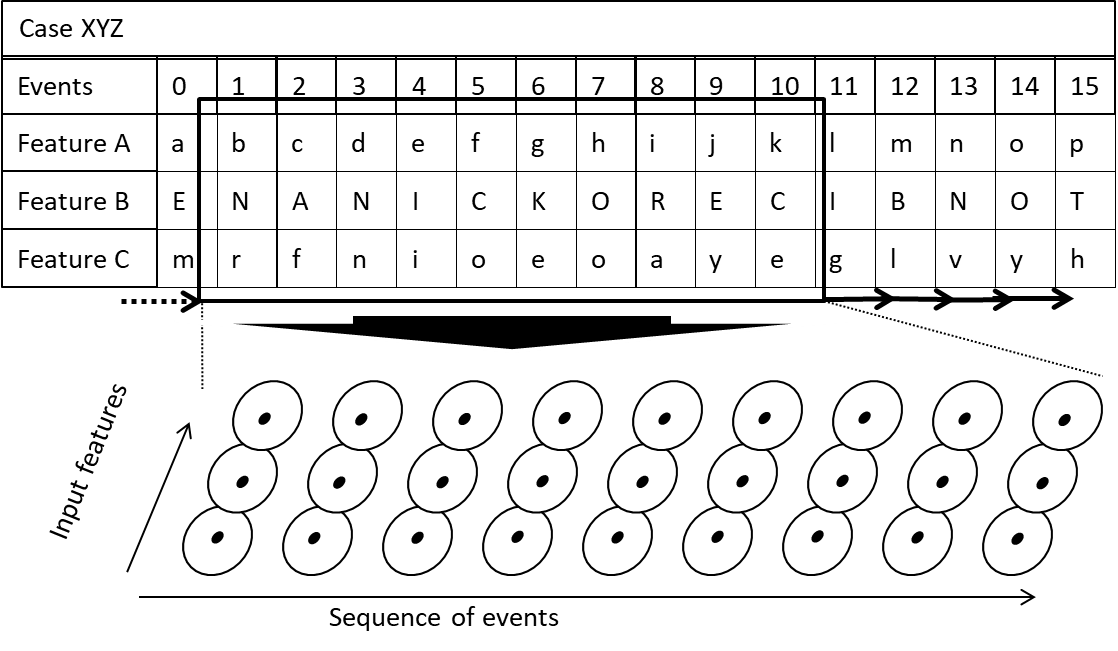}
\vspace*{-2mm}
\caption{A window with (sequence) length nine sliding over a sixteen-event case forms the input matrices for the model.} \label{fig:sequence}
\end{figure}

Since both types of LSTM learners require time sequences as inputs, we combined the vectors containing the features for all events in a given sequence into matrices with shape (sequence length x number of features). This was done by sliding a sequence-length-wide window over the events in a case. The first window only covered the first event in the case, the second one covered two, etc. The last window covered the case's sequence-length last number of events. Fig~\ref{fig:sequence} visualizes this step, showing the matrix formed by the vectors associated with events 1-10. 

\vspace*{-3mm}
Padding was used to fill in the missing values for short cases and for the first (sequence-length - 1) windows of a case. Despite their one-dimensional character suggesting otherwise, our CNNs work with the same two-dimensional input data. The one-dimensionality refers to the filters covering the whole width (number of features) and striding in one direction only, along the longitudinal axis of the matrix (sequence length). In this fashion, every case produced a number of matrices in the processed dataset equal to its number of events.
\vspace*{-1mm}

\section{Experimental evaluation} \label{sec: eval}
In this section, we first present the datasets we chose for  our experiments. We then describe the implementation of the experiments, including splitting the datasets into training, validation and test sets, hyperparameter tuning, measures against overfitting and physical infrastructure. Before finally describing our results, we devote a short paragraph to the metrics used. 

 \vspace*{-2mm}
\subsection{Datasets}\label{ss:datasets}
We selected a number of publicly available and widely-used datasets\footnote{All datasets can be found at https://data.4tu.nl/repository/collection:event\_logs\_real (4TU Centre for Research Data)} containing both range and categorical features. To discover the suitability of deep learning models in different environments, we created variety by including small and large datasets, with short and long traces, varying degrees of class balance and of lower and higher quality:

\vspace*{-1mm}
\begin{itemize}
\item \textbf{BPIC\_2012}: This dataset describes a loan application process at a Dutch bank. Every case has three possible targets: `approved', `declined' or `canceled'. This multi-classification problem was transformed into three different binary classification problems for our purposes.
\item \textbf{BPIC\_2017}: This dataset is a higher-quality version of BPIC\_2012 with both more examples and features that should facilitate better predictions. Note that the sum of the three targets exceeds 100\% as sometimes `canceled' cases are restarted and `approved' or `declined' cases become `canceled' later.
\item \textbf{Traffic fines}: Is an event log of  a system managing traffic road fines. Fines are either paid in full or sent for credit collection. The latter is our target (`deviant'). Cases in this dataset are very short (avg. 3.3 events).
\item \textbf{Sepsis cases}: Describes the pathways of patients through a hospital. We define three targets: 
	\begin{itemize}
	\item 28\_days\_EM: is the patient admitted to the emergency rooms within twenty eight days of his/her release from the hospital?
	\item IC: does the patient enter the intensive care unit?
	\item no\_A\_release: is the patient eventually released from the hospital for another reason than the most frequent `A'?
	\end{itemize}
\end{itemize}
\vspace*{-2mm}

Table~\ref{tab:data} provides an overview of the used datasets, some of which contain highly imbalanced classes. In line with our decision to apply an identical methodology to all datasets, we decided against sampling techniques to restore balance. This negatively impacted results, especially for BPIC\_2017 (declined) and Sepsis.

\begin{table}
\centering
\vspace*{-1mm}
\caption{Statistics of the used datasets.}\label{tab:data}
\scalebox{0.9}{
\begin{tabular}{|l|l|l|l|l|l|l|l|l|l|l|l|}
\hline
Dataset &  Target & \#      & \#      & Min     & Max    & Mean  & Med & Pos    & Pos     & \# categ.   & \# range\\
           &  (binary)           &events & cases & events & events & events & events & events & cases & features     & features\\
\hline
      BPIC\_            & approved  & 219,858 & 12,688 & 2 & 172 & 17.3 & 8 & .40 & .18 & 3 & 2 \\
2012 & declined  & 219,858 &12,688 & 2 & 172 & 17.3 & 8 & .26 & .60 & 3 & 2 \\
                  & canceled  &219,858 & 12,688 & 2 & 172 & 17.3 & 8 & .33 & .22 & 3 & 2 \\
\hline
    BPIC\_              & approved  & 1,071,054 & 31,417 & 7 & 175 & 34.1 & 30 & .65 & .55 & 11 & 6 \\
2017                    & declined    & 1,071,054 & 31,417 & 7 & 175 & 34.1 & 30 & .12 & .12 & 11 & 6 \\
                            & canceled  & 1,071,054& 31,417 & 7 & 175 & 34.1 & 30 & .44 & .50 & 11 & 6 \\
\hline
Traffic         & deviant               & 496,067 & 149,958 & 1 & 20 & 3.3 & 4 & .48 & .39 & 8 & 4 \\
\hline
Sepsis         & 28\_days...  & 13,095 & 781 & 5 & 185 & 16.8 & 14 & .16 & .14 & 25 & 4 \\
Sepsis         & IC                   & 10,841 & 781 & 3 & 60 & 13.9 & 13 & .09 & .14 & 25 & 4 \\
Sepsis         & no\_A...  & 13,182 & 781 & 5 & 183 & 16.9 & 14 & .16 & .14 & 25 & 4 \\
\hline
\end{tabular}
}
\vspace*{-1mm}
\end{table}

 \vspace*{-1mm}
\subsection{Implementation}
\label{subsec:impl}
We first preprocessed the data and reshaped it to fit our models as described in \ref{ss:pp} and \ref{ss:fm}.  A test set comprising the chronologically last 20\% cases of each dataset was set apart for the final evaluation of the models' predictions. We ran every model/dataset combination 50 times with different values for three hyperparameters: sequence length, batch size and model size. The latter was accomplished by multiplying every layer's number of nodes by a multiplication factor. As the CNN models had one additional hyperparameter, kernel size, they were run 100 times. Prior to every run, the input data was reshuffled before separating a training set of 80\% from a validation set of 20\% of the examples. To avoid overfitting, we used an automatic stopping mechanism to halt training after five epochs without improvement of the metric on the validation set. Each experiment was run on the Google Cloud. Polyaxon running above Kubernetes allowed for parallel execution on multiple two-core, 13GB,  2.0 GHz Intel Xeon Scalable Processors (Skylake) and either NVIDIA Tesla K80 or P100 (for BPIC\_2017) GPUs.

 \vspace*{-1mm}
\subsection{Metrics}
Given the class imbalance in some of our datasets and the binary nature of the targets, we opted for the area under the curve ROC (AUC\_ROC) metric to evaluate our models. Since AUC\_ROC yields a non-differentiable loss curve, we had to resort to accuracy to train our models. At the end of each epoch, we computed the AUC\_ROC on the validation set for the early-stopping mechanism. After training, the model with the highest AUC\_ROC score on the validation set was withheld to make predictions on the test set, on which we report below in this paper. We also calculated other metrics: F1-score, accuracy and AUC\_PR (area under the precision-recall curve) and recorded computation times.

 \vspace*{-1mm}
\subsection{Results}
The evaluation of our experiments' results allows us to formulate answers to the following questions: Which models perform better on what kind of datasets? What are the speed differences between the models? How robust are they with respect to their hyperparameters? Are early runtime predictions possible? We also make an observation about the relevance of timestamps.
 \vspace*{-1mm}
\subsubsection{CNNs perform comparably to LSTMs with and without Attention.}
Table~\ref{tab:results} shows that neural networks can deliver useful results for large enough datasets (BPIC\_2017, Traffic and BPIC\_2012). However, none of our models could cope with the combined challenge of short, imbalanced datasets and many (sparse and possibly correlated) variables found in Sepsis. Whilst the BPIC\_2012 predictions still beckon a great deal of caution (the F1-scores are extremely low for `canceled'), the BPIC\_2017 results are vastly superior and clearly demonstrate the benefits of improved data collection by the user. The lower F1- and AUC\_PR-scores for `declined' versus `approved' and `canceled' in all likelihood result from the dataset's class imbalance. Despite the extreme shortness of the cases in the Traffic dataset, the results are significantly better than random guesses. 
The three different models deployed performed equally well, suggesting they all manage to extract the same information from the data. CNNs proved to be a match for the state-of-the-art LSTM, confirming the findings of \cite{tsc, bai}. In our setting, no benefit was derived by adding an attention layer to the LSTM models.

\begin{table}[h]
\centering
\vspace*{-1mm}
\caption{Results from best models in hyperparameter space, aggregated over all prefix lengths. `Rel. Time' is the relative run time compared to the fastest run time for the respective experiment which is set at 100\%.}\label{tab:results}
\scalebox{0.82}{
\begin{tabular}{|p{1.7cm}|@{}l@{}|l@{}|}\hline
Dataset & 
\begin{tabular}{p{1.8cm}|@{}|l|@{}}
Target &
\begin{tabular} {p{1cm}|p{1cm}|p{1cm}|p{1cm}|p{1cm}|p{1cm}|p{1cm}|p{1cm}|p{1cm}|p{1cm}}
Model      & AUC\_     & F1-     & Accu-    & AUC\_  & Rel. & Batch & Seq. & Kernel & Model\\
              & ROC        &  score       &  racy      & PR        & Time & Size & Length & Size & Size\\
\end{tabular}
\end{tabular}

\tabularnewline\hline

BPIC\_2012 &
\begin{tabular}{p{1.8cm}|@{}|l|@{}}

approved &
\begin{tabular} {p{1cm}|p{1cm}|p{1cm}|p{1cm}|p{1cm}|p{1cm}|p{1cm}|p{1cm}|p{1cm}|p{1cm}}
LSTM&0.79&0.67&0.72&0.74&668\%&512&45&&16\\
Att&0.79&0.66&0.72&0.74&552\%&512&35&&16\\
CNN& \textbf{0.80}&0.69&0.72&0.75&\textbf{100\%}&256&15&8&16\\
\end{tabular}
\tabularnewline\hline

declined &
\begin{tabular} {p{1cm}|p{1cm}|p{1cm}|p{1cm}|p{1cm}|p{1cm}|p{1cm}|p{1cm}|p{1cm}|p{1cm}}
LSTM&\textbf{0.76}&0.59&0.76&0.63&443\%&1024&35&&16\\
Att&0.75&0.59&0.75&0.63&160\%&512&5&&8\\ 
CNN&\textbf{0.76}&0.59&0.75&0.62&\textbf{100\%}&512&35&4&8\\
\end{tabular}
\tabularnewline\hline

canceled &
\begin{tabular} {p{1cm}|p{1cm}|p{1cm}|p{1cm}|p{1cm}|p{1cm}|p{1cm}|p{1cm}|p{1cm}|p{1cm}}
LSTM&\textbf{0.75}&0.35&0.79&0.50&1158\%&128&5&&2\\ 
Att&\textbf{0.75}&0.35&0.79&0.50&1352\%&128&5&&4\\ 
CNN&0.74&0.26&0.79&0.48&\textbf{100\%}&1024&5&2&16\\ 
\end{tabular}

\end{tabular}
\tabularnewline\hline

BPIC\_2017 &
\begin{tabular}{p{1.8cm}|@{}|l|@{}}

approved &
\begin{tabular} {p{1cm}|p{1cm}|p{1cm}|p{1cm}|p{1cm}|p{1cm}|p{1cm}|p{1cm}|p{1cm}|p{1cm}}
LSTM&\textbf{0.93}&0.88&0.83&0.96&909\%&128&47&&4\\
Att&\textbf{0.93}&0.88&0.84&0.97&946\%&128&47&&8\\
CNN&\textbf{0.93}&0.86&0.83&0.97&\textbf{100\%}&128&47&2&4\\
\end{tabular}
\tabularnewline\hline

declined &
\begin{tabular} {p{1cm}|p{1cm}|p{1cm}|p{1cm}|p{1cm}|p{1cm}|p{1cm}|p{1cm}|p{1cm}|p{1cm}}
LSTM&\textbf{0.91}&0.66&0.92&0.69&200\%&1024&47&&16\\
Att&\textbf{0.91}&0.66&0.92&0.67&240\%&1024&47&&2\\ 
CNN&\textbf{0.91}&0.67&0.92&0.69&\textbf{100\%}&1024&47&2&4\\
\end{tabular}
\tabularnewline\hline

canceled &
\begin{tabular} {p{1cm}|p{1cm}|p{1cm}|p{1cm}|p{1cm}|p{1cm}|p{1cm}|p{1cm}|p{1cm}|p{1cm}}
LSTM&\textbf{0.92}&0.79&0.82&0.91&294\%&1024&47&&1\\ 
Att&\textbf{0.92}&0.77&0.82&0.91&356\%&512&47&&4\\ 
CNN&\textbf{0.92}&0.80&0.82&0.91&\textbf{100\%}&256&47&8&4\\ 
\end{tabular}

\end{tabular}
\tabularnewline\hline

Traffic &
\begin{tabular}{p{1.8cm}|@{}|l|@{}}

deviant &
\begin{tabular} {p{1cm}|p{1cm}|p{1cm}|p{1cm}|p{1cm}|p{1cm}|p{1cm}|p{1cm}|p{1cm}|p{1cm}}
LSTM&0.75&0.66&0.66&0.63&170\%&512&8&&2\\
Att&\textbf{0.77}&0.67&0.68&0.66&\textbf{100\%}&512&6&&2\\
CNN&0.76&0.68&0.69&0.64&132\%&128&6&3&2\\
\end{tabular}

\end{tabular}
\tabularnewline\hline

Sepsis &
\begin{tabular}{p{1.8cm}|@{}|l|@{}}

28\_days\_EM  &
\begin{tabular} {p{1cm}|p{1cm}|p{1cm}|p{1cm}|p{1cm}|p{1cm}|p{1cm}|p{1cm}|p{1cm}|p{1cm}}
LSTM&0.51&0.03&0.84&0.10&845\%&256&45&&1\\
Att&0.50&0.07&0.82&0.10&424\%&128&5&&2\\
CNN&0.47&    &0.89&0.11&\textbf{100\%}&512&25&4&4\\
\end{tabular}
\tabularnewline\hline

IC &
\begin{tabular} {p{1cm}|p{1cm}|p{1cm}|p{1cm}|p{1cm}|p{1cm}|p{1cm}|p{1cm}|p{1cm}|p{1cm}}
LSTM&0.65&0.29&0.89&0.31&191\%&256&45&&4\\
Att&0.64&0.34&0.89&0.34&125\%&256&15&&16\\ 
CNN&\textbf{0.69}&    &0.90&0.16&\textbf{100\%}&512&45&2&8\\
\end{tabular}
\tabularnewline\hline

no\_A\_release &
\begin{tabular} {p{1cm}|p{1cm}|p{1cm}|p{1cm}|p{1cm}|p{1cm}|p{1cm}|p{1cm}|p{1cm}|p{1cm}}
LSTM&0.55&0.34&0.79&0.32&447\%&128&45&&4\\ 
Att&0.56&0.27&0.78&0.32&\textbf{100\%}&256&5&&4\\ 
CNN&0.57&0.30&0.79&0.34&103\%&128&5&2&2\\ 
\end{tabular}

\end{tabular}
\tabularnewline\hline

\end{tabular}
}
\vspace*{-5mm}
\end{table}

 \vspace*{-2mm}
\subsubsection{CNNs train much faster than LSTMs.}
Training times matter, especially since concept drift will require frequent training of models in production. CNN models remarkably outrun the other models during training---sometimes by over one order of magnitude---making them the model of choice for practitioners. Prediction times are negligible. Training times depend heavily on model architecture decisions, and hardware infrastructure, and are stochastic by nature. In our implementation as described in subsection~\ref{subsec:impl}, the run times of the fastest models were 23-60, 380-500, 270, and 17-40 seconds for BPIC\_2012, BPIC\_2017, Traffic and Sepsis respectively (BPIC\_2017 was run on faster GPUs). Hyperparameter tuning  multiplies total training times by the number of trainings performed.

 \vspace*{-2mm}
\subsubsection{All models are robust with respect to their hyperparameters.}
\begin{figure}[!tbp]
\vspace*{-1mm}
  \centering

    \includegraphics[width=.8\textwidth]{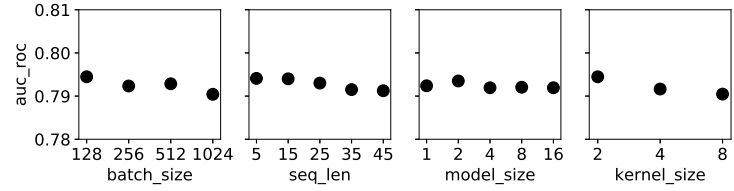}
     \vspace*{-1mm}
\caption{Hyperparameter tuning (BPIC\_2012 approved, CNN, average values for all relevant samples in 100 experiments).}
 \label{fig:hpt}
 \vspace*{+9mm}
\end{figure}

As for the hyperparameters, there is no consensus on the batch sizes: all available sizes (128, 256, 512, 1024) are used by the best models. One could expect longer sequence lengths to support better predictions for datasets with longer cases. Indeed, the sequence length is consistently about 1.5 times the median number events per case for the BPIC\_2017 and Traffic datasets. This is less clear for BPIC\_2012. We used a multiplication factor (between one and sixteen) to generate models of different widths. A preference for a certain size cannot be deduced from our experiments. The same holds true for the CNNs' kernel size.  Fortunately, these hyperparameter differences do not oblige the practitioner to engage in very extensive hyperparameter tuning. The models are robust with respect to them as visualized in Fig~\ref{fig:hpt}, where the AUC\_ROC values hardly move with changes in the individual hyperparameters.

 \vspace*{-2mm}
\subsubsection{Early runtime predictions are possible.}
\begin{figure}[]
\centering
\vspace*{-.0cm}
  \begin{tabular}{@{}c@{}} 
   \includegraphics[width=.97\textwidth]{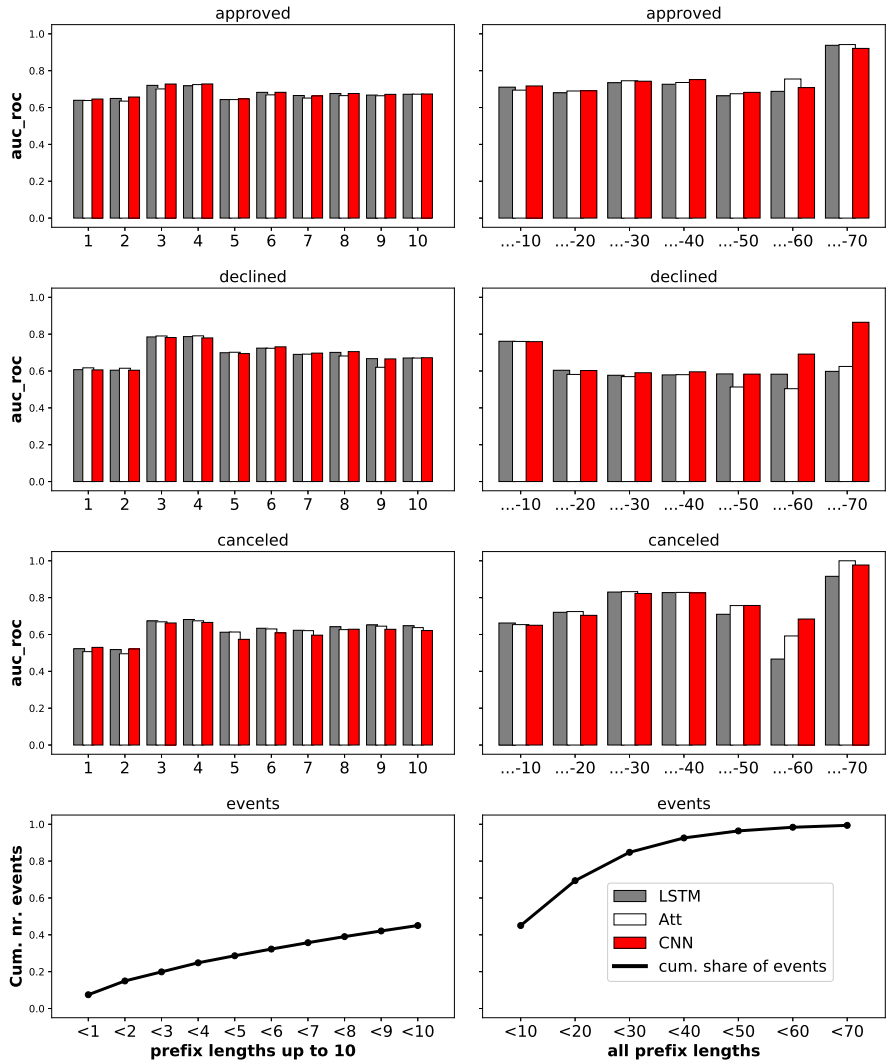}
  \end{tabular}
\vspace*{-1mm}
  \caption{Runtime predictons: AUC\_ROC as a function of prefix length (nr. of events since start of case considered) for 3 BPIC\_2012 datasets (max. prefix length of 70).}\label{fig:nrevents}
\vspace*{-.9cm}
\end{figure}

Fig~\ref{fig:nrevents} provides insight into the earliness of learning based on the example of the BPIC\_2012 dataset. The test set was sorted according to the prefix length of the (incomplete) cases. When basing predictions on up to the ten very first events of cases as in the left column of the figure, a pattern emerges: the outcomes predictions for all models are best for prefix lengths of three and four! Thus, early runtime predictions are possible as most of the information about case outcomes resides in the first few events. Waiting for more events to materialize does not seem to pay off for BPIC\_2012, as shown in the second column of Fig~\ref{fig:nrevents} where up to 70 events are considered\footnote{Notice that the (weighted) average of the AUC\_ROC of subsets will not necessarily match the AUC\_ROC of the total set, as observed here.}.

 \vspace*{-2mm}
\subsubsection{Do timestamps matter?}
\begin{figure}[!tbp]
\centering
\vspace*{-1mm}
  \begin{tabular}{@{}c@{}} 
   \includegraphics[width=0.94\textwidth]{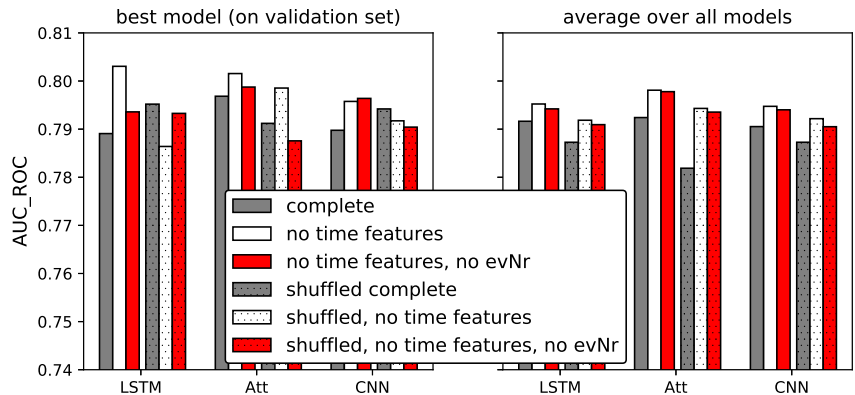}
  \end{tabular}
\vspace*{-.3cm}
  \caption{Time-related features and events order and BPIC\_2012 (approved) predictions.}\label{fig:timing}
  \vspace*{-.3cm}
\end{figure}

The aforementioned observations, combined with the CNNs' strong performance and the robustness with regards to sequence length as described earlier, lead to the conclusion that the characteristic timing feature of process mining problems does not play an important role in outcome prediction, at least not for the BPIC\_2012 dataset (approved) shown in Fig~\ref{fig:timing}.  The left-hand graph shows the results for the models that scored best on the validation sets as before. For more general conclusions, we enlarged the sample to all models trained in the hyperparameter space on the right side. The results without timestamp-related features always mildly outperform the base case (white better than grey). The effects of dropping evNr (the feature indicating the event's order in the case, red bars) and of shuffling (dotted bars) are not statistically significant. Apparently, the models are not learning from timestamp-related data, all information within the datasets is stored in the other features, an observation also made by \cite{teinemaa}.

\section{Conclusion and future work}
\vspace*{-2mm}
We found neural networks to be a useful tool for process outcome prediction, given sufficiently large datasets. CNNs deliver the same results as the state-of-the-art LSTMs at a fraction of the time and can therefore be recommended as first choice for practitioners. We found no benefit in the use of the Attention mechanism in the LSTM models. The comparison of BPIC\_2017 and BPIC\_2012 clearly demonstrates how organizations can benefit from improved data collection. Based on further exploration of the results on the BPIC\_2012 dataset, neural networks turned out to be robust with regards to their hyperparameters as well. The models often nearly reached full predictability after observing only the first few events of a case, suggesting that events critical to determining the case outcome often appear early. These factors support the usability of the models in practice, both for completed and (young) ongoing cases. The aim of this paper was to compare methods; we did not seek optimal predictions. Therefore, the published results would probably improve from balancing classes in the datasets and applying some domain-specific knowledge among other things.\\
Our conclusions on the models' performance and speed were based on experiments on several datasets with consistent results, and hence generalizable. However, the BPIC\_2012 dataset is not necessarily representative of all process mining datasets. Therefore, future research considering a wider range of datasets could solidify our conclusions about hyperparameters, runtime predictions and timestamps. We expect most of our conclusions to be valid for next-event and duration prediction problems, as the approach and models used would be identical, but only further experiments could confirm that. The observed irrelevance of the timestamp-related features warrants further inquiry. Our preliminary research suggests that under the right circumstances, neural networks outperform the classic methods in~\cite{teinemaa} to the order of a few percentage points on average. Further investigation is required to strengthen these findings and derive recommendations for the deployment of the networks. Offering insights into how the neural networks' results were reached would contribute to the practitioner's confidence in them. Finally, one could gain deeper insight into the uncertainty of the model predictions (e.g., by using Bayesian techniques), to account for the probabilistic nature of the input data and model weights.

 \vspace*{-1mm}

\end{document}